\documentclass{article}
\usepackage{spconf,amsmath,graphicx}

\usepackage{enumitem}
\usepackage{multirow}    

\setlist{nosep, leftmargin=14pt}

\usepackage{mwe} % to get dummy images

\title{LesionTABE: Equitable AI for Skin Lesion Detection}
 \name{Rocio Mexia Diaz$^{1,2}$ \qquad Yasmin Greenway$^{2}$ \qquad Petru Manescu$^{1}$}
 
 \address{$^{1}$ UCL Department of Computer Science, $^{2}$ DermieAI}
\begin{document}
%\ninept
%
\maketitle
\begin{abstract}
Bias remains a major barrier to the clinical adoption of AI in dermatology, as diagnostic models underperform on darker skin tones. We present LesionTABE, a fairness-centric framework that couples adversarial debiasing with dermatology-specific foundation model embeddings. Evaluated across multiple datasets covering both malignant and inflammatory conditions, LesionTABE achieves over a 25\% improvement in fairness metrics compared to a ResNet-152 baseline, outperforming existing debiasing methods while simultaneously enhancing overall diagnostic accuracy. These results highlight the potential of foundation model debiasing as a step towards equitable clinical AI adoption.

\end{abstract}
\section{Introduction}
\label{sec:intro}
The rising demand for dermatology services has increasingly shifted initial triage responsibilities to general practitioners (GPs). Despite being the first point of contact, GPs receive limited dermatological training, resulting in an average diagnostic accuracy of only 42\%~\cite{Tran2005DermatologyGPvsDerm}. This inaccuracy contributes to treatment delays, disease progression and by consequence, higher healthcare costs. Most consults contributing to this burden involve chronic inflammatory conditions, particularly dermatitis (eczema)~\cite{Tran2005DermatologyGPvsDerm}, which is frequently misdiagnosed as psoriasis due to their similar presentation~\cite{AboTabik2022}.

Deep learning models for skin lesion image analysis offer a promising pathway to improve triage accuracy and support chronic condition management in primary care. However, their clinical adoption remains limited by algorithmic bias caused by the under-representation of darker skin tones in training datasets~\cite{Kinyanjui2019}, leading to poorer performance for darker skin tones and widening existing inequities in dermatology care. Efforts to evaluate and mitigate such bias are constrained by the scarcity of skin tone–labelled datasets~\cite{Kinyanjui2019} and the unreliability of automated tone-labelling methods~\cite{Kalb2023}. Moreover, models demonstrating high performance focus narrowly on melanoma detection from dermatoscopic images~\cite{codella2016deeplearningensemblesmelanoma}, which does not reflect the diversity of real-world clinical dermatology.

Bias mitigation strategies can be grouped into three categories: sample balancing, synthetic data generation, and bias unlearning. Sample balancing techniques equalise tone representation but fail to eliminate disparities~\cite{Pope2024}. Synthetic data generation risks introducing artefacts or amplifying bias through skin tone domain translation~\cite{Jindal2024}. In contrast bias unlearning approaches, which remove tone-related information while preserving lesion-relevant features, have shown greater potential. For example, Bevan et al.~\cite{Bevan2021} proposed Turning A Blind Eye (TABE) using adversarial and confusion losses; Pundhir et al.~\cite{Pundhir2024} developed a variational autoencoder with adaptive latent-space resampling; and Du et al.~\cite{Du2022} introduced FairDisCo, combining adversarial and contrastive objectives. Nonetheless, these approaches employ inconsistent fairness metrics and mostly lack external validation, which limits their comparability and generalisability.

To address these gaps, this study applies a unified statistical fairness framework with external validation to advance bias mitigation in dermatological AI. It extends fairness evaluation beyond malignant lesion to inflammatory condition (eczema and psoriasis) classification using patient-captured images, and integrates a dermatology-specific foundation model to enhance representation learning. Through the integration of foundation model embeddings with adversarial debiasing, we developed LesionTABE, a novel architecture that achieves improved fairness and diagnostic performance over existing approaches.

\section{Methods}  

\subsection{Datasets} 
We used three publicly available, non-dermatoscopic, skin tone–labelled image datasets: Fitzpatrick17k~\cite{groh2022towards}, PAD-UFES~\cite{PACHECO2020106221}, and SCIN~\cite{10.1001/jamanetworkopen.2024.46615}. Over 200 diagnoses were grouped into malignant, benign, and non-neoplastic lesions, with only eczema and psoriasis retained from the latter. Fitzpatrick17k served as the training dataset for both tasks, while PADUFES was used exclusively for cancer detection testing and SCIN for inflammatory condition classification testing. The skin-tone distribution presented in Table~\ref{tab:condition_counts} shows a pronounced bias towards lighter skin tones (Fitzpatrick types I–III), representing about 75\% of all samples.

Images were resized to 224 × 224 pixels. Training augmentations included random rotations, shear transformations, horizontal and vertical flips, and brightness adjustments to improve model generalisation.
   
\begin{table}[htbp]
    \centering
    \caption{Diagnostic class counts by skin tone with totals}
    \label{tab:condition_counts}
     \setlength{\tabcolsep}{2.5pt}
    \begin{tabular}{l l c c c c c c c}
        \hline
        \textbf{Dataset} & \textbf{Lesion} & \multicolumn{6}{c}{\textbf{Fitzpatrick Skin Type}} & \multirow{2}{*}{\textbf{Total}} \\
        \cline{3-8}
        & & \textbf{I} & \textbf{II} & \textbf{III} & \textbf{IV} & \textbf{V} & \textbf{VI} & \\
        \hline
        \multirow{4}{*}{\textbf{Fitzpatrick17k}} & Malignant & 453 & 742 & 456 & 301 & 146 & 60 & \textbf{2158} \\
        & Benign & 442 & 671 & 475 & 367 & 159 & 44 & \textbf{2158} \\
        \cline{2-9}
        & Eczema & 96 & 199 & 97 & 76 & 37 & 27 & \textbf{532} \\
        & Psoriasis & 129 & 250 & 106 & 98 & 67 & 24 & \textbf{674} \\
        \hline
        \multirow{2}{*}{\textbf{PADUFES}} & Malignant & 120 & 663 & 272 & 29 & 5 & 1 & \textbf{1090} \\
        & Benign & 8 & 57 & 31 & 23 & 2 & 1 & \textbf{122} \\
        \hline
        \multirow{2}{*}{\textbf{SCIN}} & Eczema & 42 & 149 & 129 & 79 & 54 & 32 & \textbf{485} \\
        & Psoriasis & 7 & 15 & 10 & 6 & 8 & 10 & \textbf{56} \\
        \hline
    \end{tabular}
\end{table}

\subsection{Models}

Three debiasing architectures were implemented: Turning A Blind Eye (TABE)~\cite{Bevan2021}, a Variational Autoencoder with adaptive resampling (VAE)~\cite{Pundhir2024}, and FairDisCo~\cite{Du2022}. A ResNet-152 backbone pre-trained on ImageNet was used as the feature extractor for consistent comparison; this also served as the baseline. Training followed original hyperparameters from their respective studies, with the baseline trained for 10 epochs (batch size = 64, Adam, learning rate = 0.001, scheduler step = 2, $\gamma$ = 0.1) using binary cross-entropy loss. Additionally, LesionCLIP~\cite{yang2024textbook}, a dermatology foundation model trained on 438K image–text pairs from ISIC and GPT-4V reports, was evaluated as an alternative feature extractor. As ISIC mainly represents lighter Fitzpatrick types (I–III)~\cite{Kinyanjui2019}, LesionCLIP may carry tone bias; its use here tests whether dermatology-specific foundation features can generalise fairness beyond a predominantly light-skin training set.

\subsection{Metrics}

    Performance was measured using balanced accuracy. Fairness was evaluated using two metrics: Equality of Opportunity (EOM) and Predictive Quality Disparity (PQD), defined in Equation~\ref{eq:fairness_metrics}, where $\text{TPR}_{i,j} = P(\hat{y}=1 \mid y=1, j=s, i=c)$, $c$ is the diagnostic class and $s$ the skin tone group.

   \begin{equation}
\text{EOM} = \frac{1}{C} \sum_{i=1}^{C} 
\frac{\min_{j \in S} \; \text{TPR}_{i,j}}{\max_{j \in S} \; \text{TPR}_{i,j}}, 
\quad \text{PQD} = \frac{\min_{j \in S} \; \text{BA}_j}{\max_{j \in S} \; \text{BA}_j},
\label{eq:fairness_metrics}
\end{equation}

EOM measures how consistently a model identifies true positives across skin tone groups. This is critical for malignant lesion detection, where unequal sensitivity can lead to missed cancer diagnoses. PQD captures the difference in balanced accuracy across skin tones. It is particularly relevant for inflammatory conditions, where uniform diagnostic accuracy across diseases supports fair clinical management.

\subsection{Experiments}  

The Fitzpatrick17k dataset was divided into training, validation, and test sets in a 70:20:10 ratio, stratified by skin tone and condition. Batches were balanced by condition to ensure that analyses focused solely on skin tone bias. 

Models were trained on Fitzpatrick17k and evaluated through internal testing on this dataset and external testing on PADUFES (for cancer detection) and SCIN (for inflammatory lesions). Experiments were conducted using both a ResNet-152 feature extractor and LesionCLIP embeddings as input features. All experiments were repeated and averaged across five different data splits. 

Model implementations, training scripts, and evaluation code are publicly available at:

https://github.com/rociomexiadiaz/DermieAI

\section{Results}

\subsection{Malignant Lesion Detection}
Table \ref{tab:fairness_cancer} presents the Equality of Opportunity (EOM) fairness scores across all models. Using the ResNet-152 feature extractor, FairDisCo achieved the highest internal fairness on the Fitzpatrick17k dataset, while the VAE underperformed relative to the baseline. In external validation on the PADUFES dataset, this trend was reversed. When the LesionCLIP feature extractor was used, TABE consistently demonstrated the highest fairness in both internal and external evaluations.

\begin{figure}[t]
    \centering
    \includegraphics[width=1\linewidth]{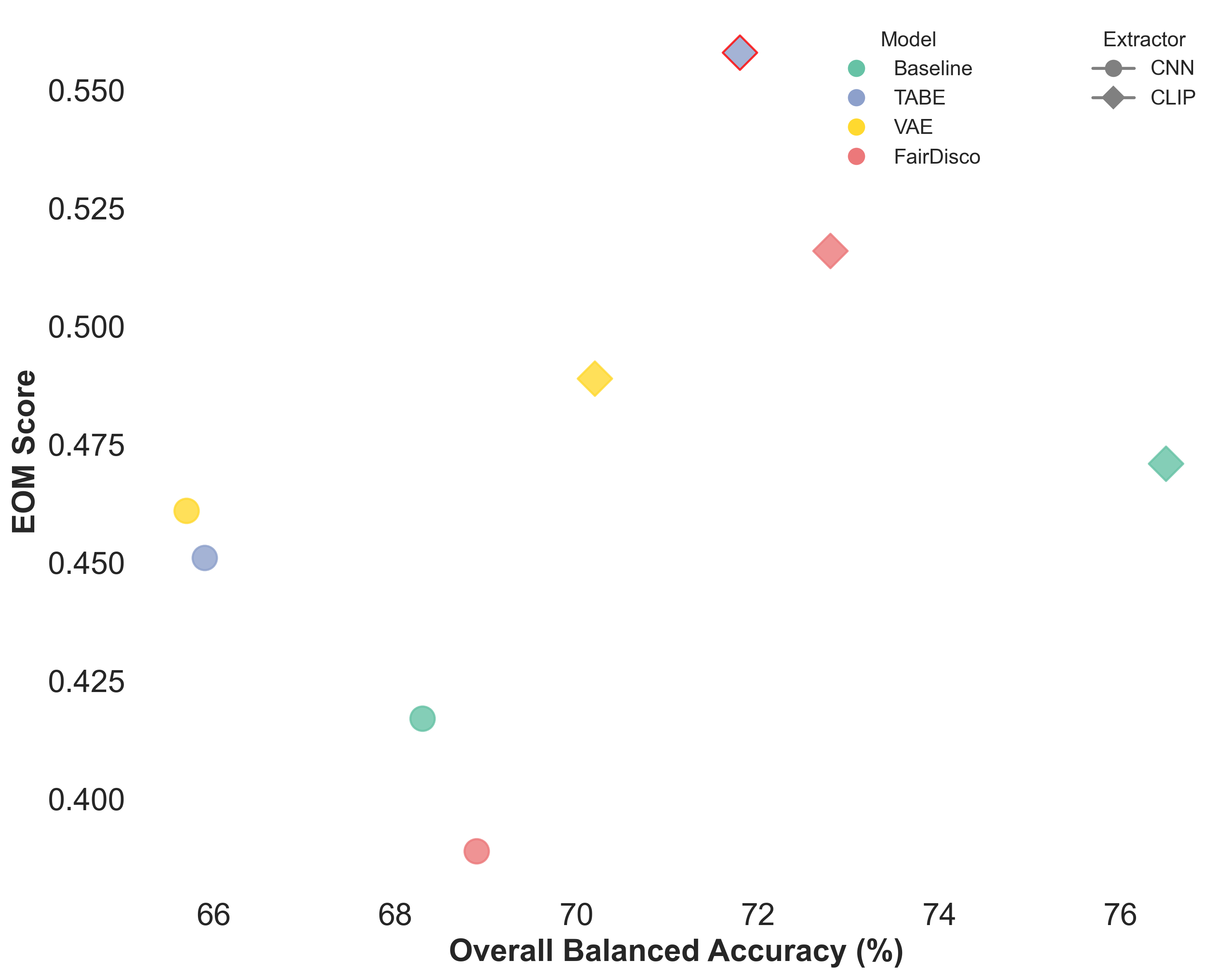}
    \caption{EOM fairness scores vs Overall Balanced Accuracy for malignant lesion detection. TABE with LesionCLIP exhibits optimal trade-off compared to all other combinations.}

    \label{fig:eom_cancer}
\end{figure}

Figure \ref{fig:eom_cancer} illustrates the trade-off between fairness and diagnostic balanced accuracy in external testing. LesionCLIP embeddings (rhombus markers) showed consistent gains in both metrics. The optimal configuration was TABE + LesionCLIP (LesionTABE) which achieved a balanced accuracy of 71.8\% and an EOM of 0.56, corresponding to relative improvements of 34\% in fairness and 5\% in performance over the baseline.

\begin{table}[htbp]
\flushleft
\caption{EOM scores for malignant lesion detection. Models were trained on Fitzpatrick17k with a ResNet-152 or LesionCLIP feature extractor and tested internally on Fitzpatrick17k, and externally on PADUFES.}
\label{tab:fairness_cancer}
\setlength{\tabcolsep}{2pt}
\begin{tabular}{lccccc}
\hline
 & \multicolumn{2}{c}{\textbf{ResNet-152}} & \multicolumn{2}{c}{\textbf{LesionCLIP}} \\
\cline{2-4} \cline{5-6}
\textbf{Model} & \textbf{Internal} & \textbf{External} &   \textbf{Internal} & \textbf{External} &  \\
\hline
Baseline  & 0.77 $\pm$ 0.11 & 0.42 $\pm$ 0.05  & 0.68 $\pm$ 0.17 & 0.47 $\pm$ 0.00\\
FairDisCo & \textbf{0.81 $\pm$ 0.07} & 0.39 $\pm$ 0.04 &  0.76 $\pm$ 0.11 & 0.52 $\pm$ 0.09 \\
TABE      & 0.79 $\pm$ 0.06 & 0.45 $\pm$ 0.17  & \textbf{0.82 $\pm$ 0.05} & \textbf{0.56 $\pm$ 0.11}  \\
VAE       & 0.77 $\pm$ 0.04 & \textbf{0.46 $\pm$ 0.10} &  0.76 $\pm$ 0.08 & 0.49 $\pm$ 0.10 \\
\hline
\end{tabular}
\end{table}

\subsection{Eczema vs Psoriasis Classification}

\begin{figure}[t]
    \centering
    \includegraphics[width=1\linewidth]{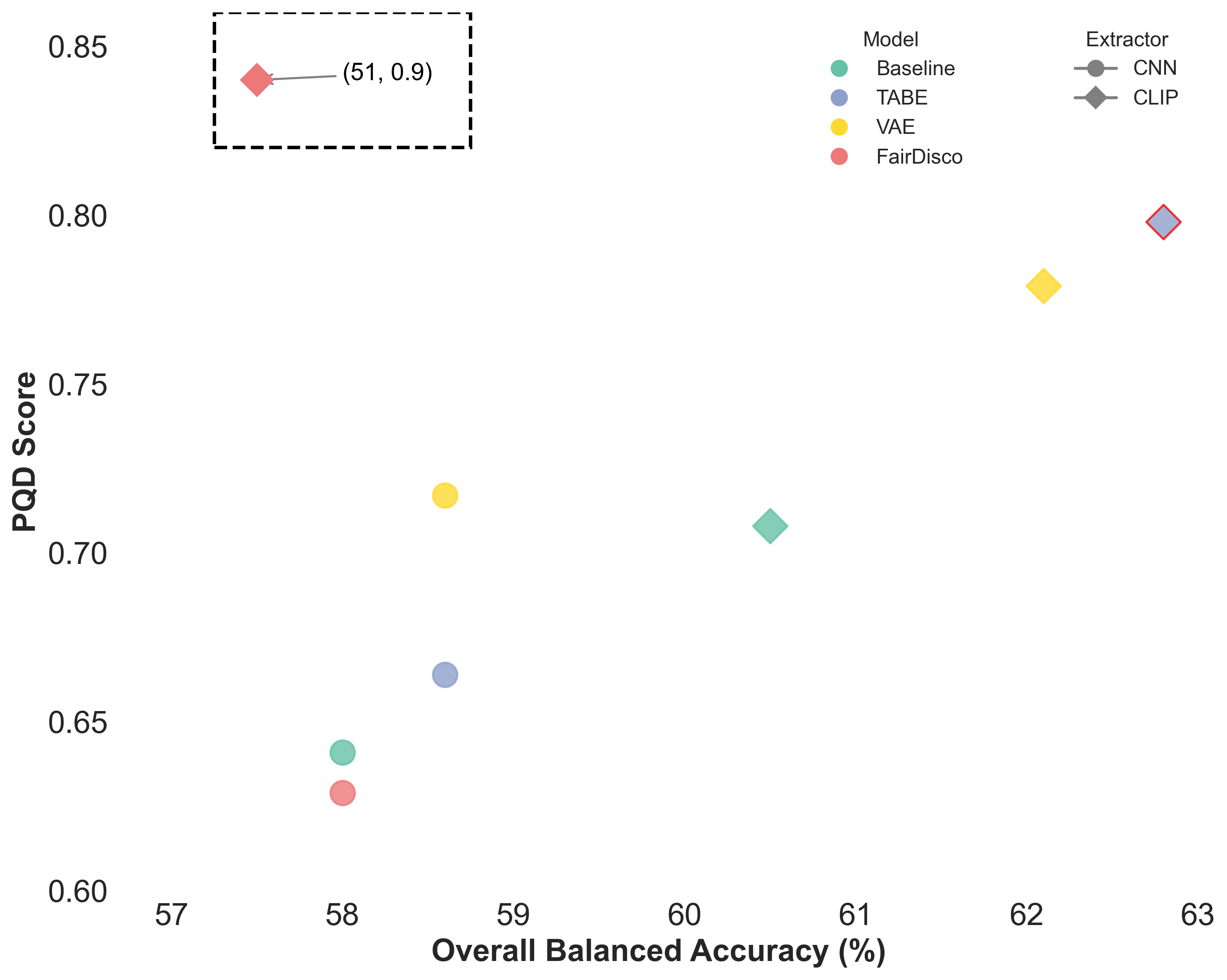}
    \caption{PQD fairness scores vs Overall Balanced Accuracy for eczema vs psoriasis classification. TABE with LesionCLIP  exhibits optimal trade-off compared to other combinations.}
    \label{fig:pqd_eczema}
\end{figure}

Table \ref{tab:fairness_EP} summarises the Predictive Quality Disparity (PQD) fairness scores across all models. Using the ResNet-152 feature extractor, none of the debiasing methods surpassed the baseline in internal evaluation on Fitzpatrick17k. In external testing on SCIN, however, the VAE achieved the highest fairness, followed closely by TABE, while FairDisCo again underperformed relative to the baseline. When the LesionCLIP feature extractor was applied, FairDisCo became the leading model for fairness in both internal and external assessments; however, this improvement came at the cost of reduced performance, as shown in Figure \ref{fig:pqd_eczema}. LesionCLIP enhanced both fairness and accuracy across most models, and the best trade-off was again achieved with LesionTABE, with a balanced accuracy of 62.8 \% and a PQD of 0.80, representing relative gains of 25\% in fairness and 8\% in balanced accuracy over the baseline.

\begin{table}[htbp]
\flushleft

\caption{PQD scores for eczema vs psoriasis classification. Models were trained on Fitzpatrick17k with a ResNet-152 or LesionCLIP feature extractor and tested internally on Fitzpatrick17k, and externally on SCIN.}
\label{tab:fairness_EP}
\setlength{\tabcolsep}{2pt}
\begin{tabular}{lccccc}
\hline
 & \multicolumn{2}{c}{\textbf{ResNet-152}} & \multicolumn{2}{c}{\textbf{LesionCLIP}} \\
\cline{2-4} \cline{5-6}
\textbf{Model} & \textbf{Internal} & \textbf{External}  & \textbf{Internal} & \textbf{External} &  \\
\hline
Baseline  & \textbf{0.73 $\pm$ 0.13} & 0.64 $\pm$ 0.10 &  0.68 $\pm$ 0.11 & 0.71 $\pm$ 0.06 \\
FairDisCo & 0.72 $\pm$ 0.08 & 0.63 $\pm$ 0.13 & \textbf{0.88 $\pm$ 0.22} & \textbf{0.92 $\pm$ 0.11}  \\
TABE      & 0.68 $\pm$ 0.08 & 0.66 $\pm$ 0.10 & 0.59 $\pm$ 0.34 & 0.80 $\pm$ 0.03  \\
VAE       & 0.62 $\pm$ 0.15 & \textbf{0.72 $\pm$ 0.07} & 0.59 $\pm$ 0.34 & 0.78 $\pm$ 0.05  \\
\hline
\end{tabular}
\end{table}

\section{Discussion}

Our novel LesionTABE framework achieved the highest fairness—surpassing the baseline by over 25\%—and delivered the most favourable balance between fairness and balanced accuracy compared to existing approaches. With diagnostic performance exceeding that of general practitioners on non-dermatoscopic, patient-captured images, LesionTABE shows strong potential for equitable real-world deployment, including AI-assisted self-triage applications. Beyond algorithmic advancements, this work introduces a unified evaluation framework with standardised fairness metrics and rigorous external validation, enabling consistent comparisons across classification tasks. Collectively, these contributions mark a significant step towards equitable AI-driven skin lesion triage systems that enhance clinical decision-making and help mitigate diagnostic disparities, particularly for malignant lesion detection, where achieving high sensitivity is crucial.

The consistent improvements in fairness and generalisation achieved with LesionCLIP embeddings provide important insights into foundation model capabilities. Despite being trained predominantly on lighter skin tones, domain-specific pretraining enables learning of lesion-relevant features that generalise more robustly across skin tone variations than generic ImageNet representations.

Finally, consistent with fairness improvements reported in the original studies using ImageNet embeddings, the VAE and TABE architectures demonstrated superior external fairness, whilst FairDisCo excelled on internal validation for malignant lesion detection. FairDisCo's marked decline on external datasets underscores the critical importance of external validation for evaluating bias mitigation strategies intended for clinical deployment.

\section{Conclusions and Future Directions}

This study showed that LesionTABE consistently enhances fairness and generalisation across diverse diagnostic tasks, outperforming existing approaches and contributing to the development of more equitable AI in dermatology.

Future work will focus on expanding LesionTABE to multi-class dermatology tasks, incorporating it into clinical triage workflows, and conducting prospective studies to evaluate its fairness, safety, and effectiveness in real-world settings.

\section{Compliance with Ethical Standards}
This study did not require ethical approval because the publicly available datasets used are de-identified and contain no personally identifiable information.

\bibliographystyle{IEEEbib}
\bibliography{strings,refs}

\end{document}